\documentclass{article}

\usepackage{jmlr2e}

\usepackage[utf8]{inputenc} 
\usepackage[T1]{fontenc}    

\usepackage{times}
\usepackage{graphicx} 
\usepackage{subfig}
\usepackage{float}
\usepackage{amsmath}
\usepackage{amssymb}
\usepackage{graphicx}
\usepackage{array}
\usepackage{tikz}
\usetikzlibrary{bayesnet}
\usepackage{algorithm}
\usepackage{algorithmic}
\usepackage{paralist}
\usepackage{pgfplots}
\usepgfplotslibrary{groupplots}
\pgfplotsset{grid style={dashed,gray}}
\usepackage{xcolor,colortbl}
\usepackage{multicol}
\usepackage{multirow}
\usetikzlibrary{shapes}
\usepackage{pgfplots}
\usepackage{pst-sigsys,pst-plot,pstricks-add}

\usetikzlibrary{external} 
\definecolor{darkblue}{rgb}{0.0, 0.0, 0.55}

\usepackage{natbib}

\usepackage{algorithm}
\usepackage{algorithmic}

\def\s{{\mathbf{s}}}

\def\x{{\mathbf{x}}}

\def\z{{\mathbf{z}}}
\def\cN{{\mathcal{N}}}
\def\y{{\mathbf{y}}}
\def\h{{\mathbf{h}}}

\def\X{{\mathbf{X}}}
\def\Y{{\mathbf{Y}}}
\def\I{{\mathbf{I}}}
\def\0{{\mathbf{0}}}

\def\E{{ \mathbb{E}}}

\def\g{{\mathbf{g}}}

\def\On{{\mathcal{O}_n}}
\def\Mn{{\mathcal{M}_n}}

\newcommand{\xhdr}[1]{\vspace{0.0mm}\noindent{{\bf #1.}}}

\makeatletter

\def\addlegendimage{\pgfplots@addlegendimage}

\begin{document}

\title{Handling Incomplete Heterogeneous Data using VAEs}

%

\author{\name Alfredo Nazabal \email anazabal@turing.ac.uk \\
\addr Alan Turing Institute\\
London, United Kingdom\\
\AND
\name Pablo M. ~Olmos \email olmos@tsc.uc3m.es\\
\addr University Carlos III in Madrid\\
Madrid, Spain\\
\AND 
\name Zoubin Ghahramani \email zoubin@eng.cam.ac.uk\\
\addr University of Cambridge\\
Cambridge, United Kingdom\\Uber AI Labs\\
San Francisco, US
\AND 
\name Isabel Valera \email ivalera@mpi-sws.org\\
\addr Max Planck Institute for Intelligent Systems\\
T\"ubingen, Germany}

\maketitle

\begin{abstract}
Variational autoencoders (VAEs), as well as other generative models, have been shown to be efficient and accurate for capturing the latent structure of vast amounts of complex high-dimensional data. However, existing VAEs can still not directly handle data that are heterogenous (mixed continuous and discrete) or incomplete (with missing data at random), which is indeed common in real-world applications. 

In this paper, we propose a general framework to design VAEs suitable for fitting incomplete heterogenous data. The proposed HI-VAE includes likelihood models for real-valued, positive real valued, interval, categorical, ordinal and count data, and allows accurate estimation (and potentially imputation) of missing data. Furthermore, HI-VAE presents competitive predictive performance in supervised tasks, outperforming supervised models when trained on incomplete data.
\end{abstract}

\section{Introduction}
Data are usually organized and stored in databases, which are often large, heterogenous, noisy, and
incomplete.
For  example, an online shopping platform has access to heterogenous and incomplete information of its users, such as their age, gender, orders, wish lists, etc. 
Similarly,  Electronic Health Records of hospitals might contain different lab measurements, diagnoses and genomic information about their patients. 
Learning generative models that accurately capture the distribution, and therefore the underlying latent structure, of such incomplete and heterogeneous datasets may allow us 
to better understand the data, estimate  missing or corrupted values, detect  outliers, and  make predictions (e.g., on patients' diagnosis) on unseen data~\citep{valera2017general}.

Deep generative models have been recently proved to be highly flexible and expressive unsupervised methods, able to capture the latent structure of complex high-dimensional data. 
They efficiently emulate complex distributions from large high-dimensional datasets,  generating new data points similar to the original real-world data, after training is completed~\citep{Kingma14,Rezende15,Li18Pami}.   So far, the main focus in the literature is to enrich the prior or posterior of explicit generative models such as variational autoencoders (VAEs); or  to propose  alternative training objectives to the log-likelihood, leading to implicit generative models such as, e.g., generative adversarial networks (GANs)~\citep{Mescheder17}. 
%
%
Indeed, we are witnessing a race between an ever-growing spectrum of VAE models, e.g., VAE with a VampPrior \citep{Tomczak18},  Output Interpretable {VAE}s  \cite{Ainsworth18}, DVAE++ \citep{vahdat2018discrete}, Shape Variational Autoencoder \citep{Nash17} and GAN-style objective functions (f-GAN \citep{f-GAN}, DR-GAN \citep{Tran18}, Wasserstein GANs \citep{WesserstainGAN},  MMD-GAN \citep{MMD-GAN}, Gated-GAN \citep{Chen19}, AdaGAN \citep{AdaGAN}, feature-matching GAN \citep{Mroueh17-2}, etc.). 
While all these approaches compete to generate the most realistic images or readable text, 
the deployment of such models to solve practically-relevant problems in arbitrary datasets, which are often incomplete and heterogenous \cite{FARHANGFAR20083692}, is being overlooked. 
{In the following, we discuss these problems in more detail and why we believe our paper is relevant to data-scientists interested in exploiting the deep generative model pipeline in the data wrangling process.  We provide with practical tools to handle both missing  and heterogeneous data with little supervision from the user, who merely has to indicate the data type model of each attribute {and the position of the missing data}.}

%
Currently deep generative models focus on  highly-structured homogeneous data collections including, e.g.,   images \citep{Salimans16,ZHAO2019356}, text  \citep{Yang17}, video \citep{Cao19,YU2019179,ATAPOURABARGHOUEI2019232}  or speech \citep{Freitag2017}, which are  characterized by strong statistical dependencies between pixels or words. The dominant existing approach to account for heterogenous data follows a deep domain-alignment approach \citep{Ganin2016,Kim17,Taigman16}, designed to discover relations between two unpaired unlabelled datasets rather than modelling their joint distribution using a probabilistic generative model \citep{Liu2017,Zhu17,castrejon2016}. 
Surprisingly, not much attention has been paid to describing how deep generative models can be designed to effectively learn the distribution of less structured, heterogeneous datasets. In these datasets there is no clear notion of correlation among the different attributes (or dimensions) to be exploited by weight sharing using convolutional or recurrent neural networks.  
{As we show in this paper, preventing a few dimensions of the data dominating the training  is a crucial aspect to effectively deploy deep generative models suitable for heterogeneous data. 

%

%
Similarly,  there is no clear discussion  in the current literature on  how to incorporate missing data during the training of deep generative models. 
Existing approaches  consider either complete data during training~\citep{Liu18}, or assume incomplete information only in one of the dimensions of the data, which corresponds to the one they aim to predict (e.g., the label in a classification task)~\citep{Kiyuk15,Kingma14-2}. {However, both approaches are not realistic enough, since it might be crucial for the performance of an unsupervised model to use all the available information during training.  
Recently, \citep{yoon2018gain} proposed a GAN approach, named as GAIN, to input missing data, where the generator completes the missing values given the observed ones, and the discriminator aims to distinguish between true and imputed values. However, this approach can only handle continuous or binary data, and it is not easily generalizable to heterogeneous data.
As a consequence,  strategies to effectively train deep generative models on incomplete and heterogeneous datasets are still required}.

In this work, we present a general framework for VAEs that effectively incorporates incomplete data and heterogenous observations. Our design presents the following features:
%
%
%
%
%
\begin{compactitem}

\item[i)] a generative model that handles mixed numerical (continuous real-valued and positive real-valued,  as well as discrete count data) and nominal (categorical and ordinal data) likelihood models, which we parametrize using deep neural networks (DNNs);
\item[ii)]  a stable recognition model that handles Missing Data Completely at Random  (MCAR) without increasing its complexity or promoting overfitting; 
\item[iii)] {a data-normalization input/output layer that prevents a few dimensions of the data dominating the training of the VAE,} improving also the training convergence; and 
\item[iv)] an ELBO (Evidence Lower Bound), used to optimize the parameters of both  the generative and the  recognition models, that is computed only on the observed data, regardless of the pattern of  missing data.

\end{compactitem} 
The resulting VAE is a fully unsupervised model which allows us not only to accurately solve unsupervised tasks, such as density estimation or missing data completion, but also supervised tasks (e.g., classification or regression ) with incomplete input data.  We provide the reader with specific guidelines  to design  VAEs for real-world data, which are  compatible with modern efforts in the design of VAEs and implicit models (GANs), mainly oriented to prevent the mode-dropping effect \citep{WesserstainGAN,Arora17-2}. Our empirical results show that our proposal outperforms competitors, including the recent GAIN~\citep{yoon2018gain},  on a heterogenous data completion task, and provides comparable accuracy in classification tasks to deep supervised  methods--which cannot handle missing values in the input data, therefore, requiring imputing missing inputs in the data.

\section{Problem statement }\label{sec:model}

We define a heterogeneous dataset as a collection of $N$ objects, where each object is defined by $D$ attributes and these attributes correspond to either numerical (continuous or discrete) or nominal variables. We denote each object in the dataset as a $D$-dimensional vector $\x_n= [x_{n1}, \ldots, x_{nD}]$, where each attribute $x_{nd}$ corresponds to one of the following data types: 
\begin{compactitem}
\item Numerical variables:
\begin{compactenum}
\item Real-valued data, which takes values in the real line, i.e., $x_{nd}\in \mathbb{R}$.
\item Positive real-valued data, which takes values in the positive real line, i.e., $x_{nd} \in \mathbb{R}^+$.
\item (Discrete) count data, which takes values in the natural numbers, i.e., $x_{nd}\in \{1, \ldots, \infty\}$.
\end{compactenum}
\item Nominal variables:
\begin{compactenum}
\item Categorical data, which takes values in a finite unordered set, e.g., $x_{nd}\in\{$`blue', `red',  `black'$\}$.
\item Ordinal data, which takes values in a finite ordered set, e.g., $x_{nd}\in\{$`never',  `sometimes', `often', `usually', `always'$\}$.
\end{compactenum}
\end{compactitem}
%

Additionally, we consider that a random set of entries in the data is incomplete, {under the MCAR assumption \citep{rubin1976inference}}, such that each object $\x_n$ can potentially correspond to any combination of 
observed and missing attributes. Let $\On$ ($\Mn$) be the index set of observed (missing) attributes for the $n$-th data point, where $\On\cap\Mn=\emptyset$. Also, let $\x^o_n$ ($\x^m_ n$) represent the sliced $\x$ vector, including only the elements indexed by $\On$ ($\Mn$). Figure~\ref{fig:model}(a) shows an example of an incomplete heterogenous dataset, where we observe that the different attributes (or dimensions) in the data correspond to different types of numerical and nominal variables, and missing values appear ubiquitously across the data. 

Diverging from common trends in the deep generative community, we consider databases that do not contain highly-structured  homogeneous data, but instead each observed object is a set of scalar mixed numerical and nominal attributes, being the correlations between attributes (the underlying structure), in many cases, weak. Since the dimensionality of these datasets can be relatively small (compared to images for instance), we need to carefully design the generative model to avoid overfitting on the observed data, while keeping the model flexible enough to incorporate both heterogeneous data types and random patterns of missing data.

\section{Generalizing  VAEs for Heterogeneous and Incomplete Data}\label{sec:model2} 

In this section, we show how to extend the vanilla VAE introduced in \citep{Kingma14} to handle incomplete and heterogeneous data. 

\begin{figure}[t!]
\centering\includegraphics[scale=0.45]{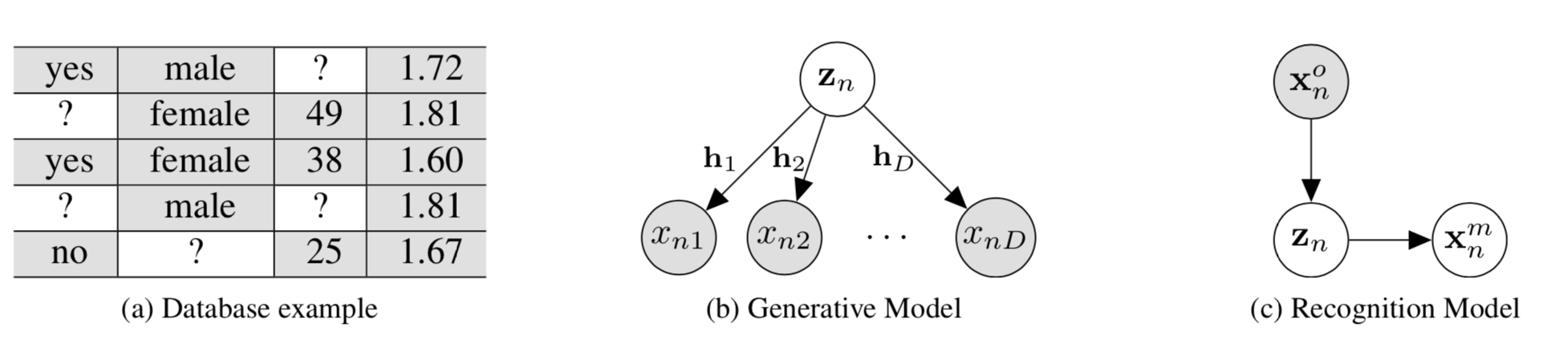}
\caption{(a) Example of incomplete heterogenous data. Panel (b)  shows our generative model, where every dimension in the observation vector $\x_n=[x_{n1}, \ldots, x_{nD}]$ corresponds to either a numerical or nominal variable, and therefore,  the likelihood parameters of each dimension $d$ are independently provided by an independent DNN $\h_d$. Additionally, panel (c) shows our recognition model to infer the missing data $\x_n^m$ from observed data $\x_n^o$.} \label{fig:model}
\end{figure}

\subsection{Handling incomplete data}\label{sec:missing}

In a standard VAE, missing  data affect both the generative (decoder) and the recognition (encoder) models. The ELBO is defined over the complete data, and it is not straightforward to decouple the missing entries from rest of the data, particularly when these entries appear completely at random in the dataset. To this end, we first propose to use the following factorization  for the decoder (Figure \ref{fig:model}(b)):
%
%
\begin{equation}\label{eq:genMsimple}
p(\x_{n},\z_n) = p(\z_n)\prod_{d}  p(x_{nd}|  \z_{n}),
\end{equation}
where $\z_n \in \mathbb{R}^K$ is the latent $K$-dimensional vector representation of the object $\x_n$, and $p(\z_n)=  \cN(\z_n|\boldsymbol{0}, \mathbf{I}_K)$. This factorization makes it easy to marginalize out the missing attributes for each object from the variational ELBO. We parametrize the likelihood $p(x_{nd}| \z_{n}) $ with the set of parameters $\boldsymbol{\gamma}_{nd}= \h_d(\z_{n})$, where $\h_d(\z_{n})$ is a DNN that transforms the latent variable $\z_n$ into the likelihood parameters $\boldsymbol{\gamma}_{nd}$. 

%
Note that the above factorization of the likelihood allows us to separate the contributions of the observed data $\x_n^o$ from the missing data $\x_n^m$ as
\begin{equation}\label{eq:genMsimple2}
p(\x_{n}|\z_n) = \prod_{d \in \On}  p(x_{nd}| \z_{n}) \prod_{d \in \Mn}  p(x_{nd}| \z_{n}).
\end{equation} 


%
The recognition model, graphically represented in Figure~\ref{fig:model}(c), also needs to account for incomplete data, such that the distribution of the latent variable $\z_n$ only depends on the observed attributes $\x_n^o$, i.e., 
\begin{equation}\label{qmodel}
q(\z_n, \x_n^m|\x_n^o ) = q(\z_n | \x_n^o) \prod_{d \in \Mn} p(x_{nd} | \z_n ).
\vspace{-5pt}
\end{equation}

Given the above generative and recognition models, described respectively by  \eqref{eq:genMsimple} and \eqref{qmodel}, the ELBO of the marginal likelihood (computed only on the observed data $\X^o$)   can be written as 
\begin{align}\label{eq:ELBO}
\vspace{-5pt}
\log p(\X^o)  &=\sum_{n=1}^{N} \log p(\x^o_n) \nonumber = \sum_{n=1}^{N} \log\int p(\x^o_n,\x^m_n,\z_n) d\z_n~d\x^m_n \nonumber\\ 
 &\geq  \sum_{n=1}^{N}  \E_{q (\z_n |\x_n^o )}  \left[\sum_{{d \in \On}} \log p(x_{nd} | \z_n) \right]-  \sum_{n=1}^{N} \mathrm{KL} \left( q(\z_n |\x_n^o ) || p(\z_n )\right),
\end{align}
where the first term of the ELBO corresponds to the reconstruction term of  (\emph{only}) the observed data $\X^o$, and the Kullback-Liebler (KL) divergence  in the second term penalizes any deviation of the posterior $q(\z_n |\x_n^o )$ from the prior $p(\z_n)$. Note that the KL divergence can be computed in closed-form~\citep{Kingma14}
\\

\xhdr{Recognition models for incomplete data}

We need an encoder that is flexible enough to handle any combination of observed and missing attributes.  
To this end, we propose an \textit{input drop-out} recognition distribution whose parameters are the output of a DNN with input $\tilde{\x}_n$, such that
\begin{align}\label{eq:qmodel2}
q(\z_n |\x_n^o )&=\cN\left(\z_n|\boldsymbol{\mu}_{q}(\tilde{\x}_n ),\boldsymbol{\Sigma}_q(\tilde{\x}_n)\right) ,
\vspace{-5pt}
\end{align}
where the  input $\tilde{\x}_n$ is a $D$-length vector that resembles the original observed vector $\x_n$ but the missing dimensions are replaced by zeros, and $\boldsymbol{\mu}_{q}(\tilde{\x}_n )$ and $\boldsymbol{\Sigma}_q(\tilde{\x}_n)$ are parametrized DNNs with input $\tilde{\x}_n$ whose output determine the mean and the diagonal covariance matrix of \eqref{eq:qmodel2}. 
In order to  make sure that the missing inputs do not affect to the output of the encoder (nor to the learning of its parameters), we need to ensure that the contribution of the missing attributes to the encoder outputs and the evaluation of the derivatives  with respect to the network parameters of $ \boldsymbol{\mu}_{q}(\tilde{\x}_n  )$ and $\boldsymbol{\Sigma}_{q}(\tilde{\x}_n )$  is zero. To this end, we rely on multilayer perceptron neural network architectures, where the output of every neuron is a non-linear transformation of a (linear) weighted sum of the inputs, and thus the output (and its derivative) does not depend on the zero entries.


{An alternative approach, proposed in \citep{vedantam2017generative}, consists of exploiting the properties of Gaussian distributions in the linear factor analysis case \citep{williams2018autoencoders} and extending them to non-linear models, designing a factorized recognition model:} 
$$q(\z_n |\x_n^o) = p(\z_n)  \prod_{d \in \On} q(\z_n | x_{nd}),$$ 
where $q(\z_n | x_{nd}) = \mathcal{N} \left(\z_n|\boldsymbol{\mu}_{d}(x_{nd} \right), \boldsymbol{\Sigma}_{d}(x_{nd} )) $, and therefore, $q(\z_n |\x_n^o  ) =\cN\left(\z_n|\boldsymbol{\mu}_{q}(\x_n^o ),\boldsymbol{\Sigma}_{q}(\x_n^o )\right)$
with
\begin{align}\label{factorized}
\boldsymbol{\Sigma}^{-1} _{q}(\x_n^o ) & =   \mathbf{I}_K + \sum_{d \in \On}\boldsymbol{\Sigma}^{-1} _{d}(x_{nd }), \hspace{7pt} \\
 \quad 
\boldsymbol{\mu}_{q}(\x_n^o ) &= \boldsymbol{\Sigma}_{q}(\x_n^o  ) \left( \sum_{d\in \On} \boldsymbol{ \mu}_{d}(x_{nd} ) \boldsymbol{\Sigma}_{d}^{-1}(x_{nd} ) \right). 
\end{align}
Note that, in contrast to our input drop-out recognition model, in this case we need to train an independent DNN per attribute $d$, which might not only result in a higher computational cost, as well as in overfitting, {but it also loses the ability of DNNs to amortize the inference of the parameters across  attributes, and therefore, across different missing data patterns. }

%

\xhdr{Remark} This VAE for incomplete data can readily be used to estimate the missing values in the data as follows
\begin{equation}\label{eq:missingEst}
p(\x_n^m | \x_n^o) \approx \int p(\x_n^m|\z_n) q(\z_n |\x_n^o ) d\z_n
\end{equation}
The KL term in \eqref{eq:ELBO},  promotes a missing-data recognition model  that does not rely on the observed attributes, i.e., $p(\x_n^m | \x_n^o)  \approx  \int p(\x_n^m|\z_n) \mathcal{N}(\z_n | \mathbf{0}, \I_K) d\z_n$.  {In those cases where the  KL term in~\eqref{eq:ELBO} tends to dominate the ELBO, we can modify the probabilistic model to favour richer structures in the posterior distribution by replacing the independent Gaussian prior with a more structured distribution such as a mixture model.} {We discuss this approach in more detail in Section~\ref{extensions}}.


%

\subsection{Handling heterogenous data}\label{sec:het}

Standard applications of VAEs  consider homogeneous data (e.g., images) where all the observed attributes (pixels) share the likelihood function (e.g., a Gaussian or a Bernoulli distribution~\cite{Kingma14}) whose parameters are  often jointly modeled by a single NN (e.g., a convolutional DNN). 
In contrast, our setting assumes that every attribute in the data may correspond to  one of the numerical or nominal  data types introduced in Section \ref{sec:model}, and thus it requires an appropriate likelihood function. 
 Assuming the factorized decoder in \eqref{eq:genMsimple}, we can easily accommodate a variety of likelihood functions, one per input attribute, where an independent DNN, $\h_d(\cdot)$, is used to determine the  parameters $\boldsymbol{\gamma}_{nd}$ of every likelihood model $p(x_{nd}| \z_{n}) = p(x_{nd}| \boldsymbol{\gamma}_{nd}=\h_d(\z_{n}))$, as shown in Figure~\ref{fig:model}(b).  
%
Next, we define suitable likelihood functions to model the numerical and nominal  data types introduced in Section \ref{sec:model}, and show how to parameterize these likelihood functions using DNNs. We remark, that while here we have selected common choice likelihood functions as showcase examples -- e.g., log-Normal, ordinal logit-function and Poisson distributions to respectively model positive real-valued, ordinal categorical nominal, and count variables --,  other distributions, such as the Gamma distribution for positive real data or the negative binomial distribution for count data, could alternatively be used.

\xhdr{1. Real-valued data}  For real-valued data, we assume a Gaussian likelihood model, i.e., 
 \begin{equation}
  p(x_{nd}| \boldsymbol{\gamma}_{nd}) = \cN\left(x_{nd}|\mu_d(\z_{n}), \sigma_d^2(\z_n)\right),
  \end{equation}
   with $\boldsymbol{\gamma}_{nd}=\{\mu_d(\z_{n}), \sigma_d^2(\z_n)\}$, where the mean $\mu_d(\z_{n})$ and the variance  $\sigma_d^2(\z_n)$ are computed as the outputs of DNNs with input $\z_n$. 


\xhdr{2. Positive real-valued data}  For positive real-valued data, we assume a log-normal likelihood model,  i.e.,  
\begin{equation}
 p(x_{nd}| \boldsymbol{\gamma}_{nd}) = \log\cN\left(x_{nd}|\mu_d(\z_{n}), \sigma_d^2(\z_n)\right),
 \end{equation}
with $\boldsymbol{\gamma}_{nd}=\{\mu_d(\z_{n}), \sigma_d^2(\z_n)\}$, where the likelihood parameters $\mu_d(\z_{n})$ and  $\sigma_d^2(\z_n)$ (which corresponds to the mean and variance of the variable's natural logarithm) are the outputs of DNNs with input $\z_n$. 

\xhdr{3. Count data}  For count data $x_{nd} \in \{{0},1,2,\ldots, \infty\}$, we assume a Poisson likelihood model, i.e, 
\begin{equation}
p(x_{nd} | \boldsymbol{\gamma}_{nd}) = \textrm{Poiss}\left(x_{nd}|\lambda_d(\z_n)\right) = \frac{(\lambda_d(\z_n))^{x_{nd}} \exp(-\lambda_d(\z_n))}{x_{nd}!},
\end{equation}
 with $\boldsymbol{\gamma}_{nd}= \lambda_d(\z_n)$, where the mean parameter of the Poisson distribution corresponds to the output of a DNN.

\xhdr{4. Categorical data}  For categorical data, codified using one-hot encoding, we assume a multinomial logit model such that the $R$-dimensional output of  a DNN $\boldsymbol{\gamma}_{nd}=\{h_{d0}(\z_n), h_{d1}(\z_n), \ldots, h_{d(R-1)}(\z_n)\}$ represents the vector of unnormalized probabilities, such that the probability of every category is given by 
\begin{equation}
p(x_{nd} = r|\boldsymbol{\gamma}_{nd}) = \frac{\exp(-h_{dr}(\z_n))}{\sum_{q=1}^{R}\exp(-h_{dq}(\z_n))}.
\end{equation}
To ensure identifiability, we fix the value of $h_{d0}(\z_n)$ to zero.

\xhdr{5. Ordinal data}  For ordinal data, codified using thermometer encoding,\footnote{As an example, in an ordinal variable with three categories the lowest value is encoded as ``100'', the middle value as ``110'' and the highest value as ``111''.} we assume the ordinal logit model~\citep{mccullagh80}, where the probability of each (ordinal) category can be computed as 
\begin{equation}
p(x_{nd} = r|\boldsymbol{\gamma}_{nd}) = p(x_{nd}\leq r|\boldsymbol{\gamma}_{nd})-  p(x_{nd}\leq r-1|\boldsymbol{\gamma}_{nd}), 
\end{equation}
with
\begin{equation}
p(x_{nd}\leq r|\z_n) = \frac{1}{1+\exp(-(\theta_r(\z_n) - {h}_{d}(\z_n)) ) }.
\end{equation}
Here, the thresholds $\theta_{r}(\z_n)$ divide the real line into $R$ regions and $\mathbf{h}_{d}(\z_n)$ indicates the region (category) in which $x_{nd}$ falls. Therefore, the likelihood parameters are $\boldsymbol{\gamma}_{nd}= \{ {h}_{d}(\z_n), \theta_1(\z_n)  \ldots, \theta_{R-1} (\z_n) \}$, which we model as the output of a DNN. To guarantee that  $\theta_{1}(\z_n) < \theta_{2}(\z_n)< \ldots < \theta_{R-1}(\z_n)$, we apply a cumulative sum function to the $R-1$ positive real-valued outputs of the network.

%
{Moreover, for all the likelihood parameters which need to be positive, we use the softplus function $f(x) = \log(1+ \exp(x))$ }. 

%

\xhdr{Remark}{ The caveat of the generative model in Figure~\ref{eq:genMsimple} is that we are losing the ability of deep neural networks  to capture correlations among data attributes by amortizing the parameters, {since we are learning a different network to link the latent variable $\mathbf{z}$ to each particular attribute by modeling the parameters of a certain observation model $p(\mathbf{x}_d|\mathbf{z})$.} An alternative would be to use the approach in \citep{suh2016gaussian}, where categorical one-hot encoded variables are  approximated by continuous variables using jitter noise (uniform on [0,1]). {When all attributes are assumed to be continuous, we could use a single network to map $\mathbf{z}$ to the parameters (mean and covariance) of a $D$-dimensional Gaussian distribution.} However, this approach does not allow a combination of different likelihood models or distinguish categorical and ordinal data.  In Section~\ref{extensions}, we show how to solve this limitation by using a hierarchical model.} 

\xhdr{Handling heterogenous data ranges}
{Apart from different types of attributes, heterogeneous datasets commonly contain numerical attributes whose values correspond to completely different domains. For example, a dataset may contain the height of different individuals with values in the order of $1.5-2.0$ meters, and  also their income, which might reach tens or even hundreds thousands of dollars per year. }
In order to learn the parameters of both the generative and the reconstruction models in Figure~\ref{fig:model}, one might  rely on stochastic gradient descent using at every iteration a minibatch estimate of the ELBO in~\eqref{eq:ELBO}.\footnote{Although here we use the standard ELBO for VAEs, tighter log-likelihood lower bound, such  as the one proposed in the importance weight encoder (IWAE) in~\citep{burda2015importance}, could also be applied.} However, the heterogenous nature of the data and these differences of value ranges between continuous variables result in broadly different likelihood parameters (e.g., the mean of the height is much lower than the mean of the income), leading in practice to heterogenous (and potentially unstable) gradient evaluations. 
To avoid the gradient evaluations of the ELBO being dominated by a subset of attributes, we apply a batch normalization layer at the input of the reconstruction model for the numerical variables, and we apply  
the complementary batch denormalization at the output layer of the generative model to denormalize the likelihood parameters.

In particular, for real-valued variables, we shift and scale the input data to the recognition model to ensure that the normalized minibatch has zero mean and variance equal to one. These shift and scale normalization parameters, $\mu'$ and $\sigma'$, are afterwards used to denormalize the likelihood parameters of the Gaussian distribution, i.e., $x_{nd}\sim \cN\left(x_{nd}|\sigma' \boldsymbol{\mu}_d(\z_n) + \mu' , \sigma'^2\boldsymbol{\sigma}^2_d(\z_n)\right)$. 
For positive real-valued variables, for which a log-Normal model is used, we apply the same batch normalization at the encoder and denormalization at the decoder used for real-valued variables, but to the natural logarithm of the data, instead of directly to the data. 
%
%
We note that count variables are not batch denormalized at the decoder, but a normalized $\log(\cdot)$ transformation is used to feed the recognition network. With this batch normalization and denormalization layers at respectively the recognition and the generative models, we do not only enforce more stable evaluations (free of numerical errors) of the gradients, but we also speed-up the convergence of the optimization.   

\begin{table}[t!]
\caption{HI-VAE probabilistic model} \label{HIVAE-table}
\begin{tabular}{ll}
\hline 
\textbf{Generative}         &$p(\x_{n},\z_n, \s_n) = p(\s_n)p(\z_n|\s_n)\prod_{d}  p(x_{nd}| \boldsymbol{\gamma}_{nd}= h_d(\y_{nd},\s_n))$, where \\
& 
where $\Y_n = [\y_{n1}, \ldots, \y_{nD} ] = \g(\z_n )$  \\ 
& \quad \quad \quad  \!\! $p(\s_n) = \text{Categorical}(\s_n|\boldsymbol{\pi})$
,\quad   \!\! $p(\z_n| \s_n) = \cN(\z_n|\boldsymbol{\mu}_p(\s_n), \mathbf{I}_K)$ \\\\
\textbf{Recognition }            & $q(\s_n,\z_n, \x_n^m|\x_n^o ) = q(\s_n| \x_n^o) q(\z_n | \x_n^o, \s_n ) \prod_{d\in \Mn} p(x_{nd} | \z_n, \s_n ),$\\& where $q(\s_n|\x_n^o) = \text{Categorical}(\s_n|\boldsymbol{\pi}(\tilde{\x}_n))$ \\& \quad \quad \quad  \!\!  $q(\z_n|\x_n^o, \s_n) = \cN(\z_n|\boldsymbol{\mu}_q(\tilde{\x}_n,\s_n), \boldsymbol{\Sigma}_q(\tilde{\x}_n,\s_n))$ \\\\
\textbf{ELBO } 	& $\log p(\X^o)  \geq  \sum_{n=1}^{N}\left(  \E_{q(\s_n, \z_n |\x_n^o )}  \left[\sum_{d\in\On} \log p(x_{nd} | \z_n, \s_n) \right] \right)$ \\
& $\qquad\qquad~~~- \sum_{n=1}^{N} \E_{q(\s_n|\x_n^o ) } \left[ KL \left( q(\z_n |\x_n^o, \s_n )|| p(\z_n | \s_n)\right)\right]$ \\&  $\qquad\qquad~~~- \sum_{n=1}^{N}  KL \left( q(\s_n |\x_n^o)|| p(\s_n)\right)$ \\\\
\textbf{Likelihoods}         & {Real-valued data (Normal):}   $\boldsymbol{\gamma}_{nd}=\{\mu_d(\y_{nd},\s_n), \sigma_d^2(\s_n)\}$ \\
 &    Positive real-valued data (log-Normal):  $\boldsymbol{\gamma}_{nd}=\{\mu_d(\y_{nd},\s_n), \sigma_d^2(\s_n)\}$ \\
  &   Count data (Poisson):   $\boldsymbol{\gamma}_{nd}=\lambda_d(\y_{nd},\s_n)$\\
  & Categorical (Mult. logit):   $\boldsymbol{\gamma}_{nd}=\{h_{d0}(\y_{nd},\s_n), \ldots, h_{d(R-1)}(\y_{nd},\s_n)\}$\\
  & Ordinal (Ordinal logit):    $\boldsymbol{\gamma}_{nd}= \{ {h}_{d}(\y_{nd},\s_n), \theta_1(\s_n)  \ldots, \theta_{R-1} (\s_n) \}$\\
\hline \\
\end{tabular}
\end{table}


\section{The Heterogeneous-Incomplete VAE (HI-VAE)}\label{extensions}

In the previous section, we have introduced a simple VAE architecture that handles incomplete and heterogeneous data. However, this approach might be too restrictive to capture complex and high-dimensional data. 
More specifically, we have assumed a standard Gaussian prior on the latent variables $\z_n$, which  might be too restrictive based on the literature~\citep{Tomczak18} and may be particularly problematic when the final goal is to estimate missing values in unstructured datasets (refer to the discussion under \eqref{eq:missingEst}). 
Similarly, we have assumed a generative model that fully factorizes for every (heterogenous) dimension in the data, losing the properties of an amortized generative model where the different dimensions share the weights of a common DNN capturing the relationships between attributes (as CNNs capture correlations between pixels in an image). 
In this section, we overcome these limitations of the model discussed in the previous section and remark that the models proposed in this paper are, in fact, compatible with the current developments in VAE literature.

\begin{figure}[t!]
\centering\includegraphics[scale=1.0]{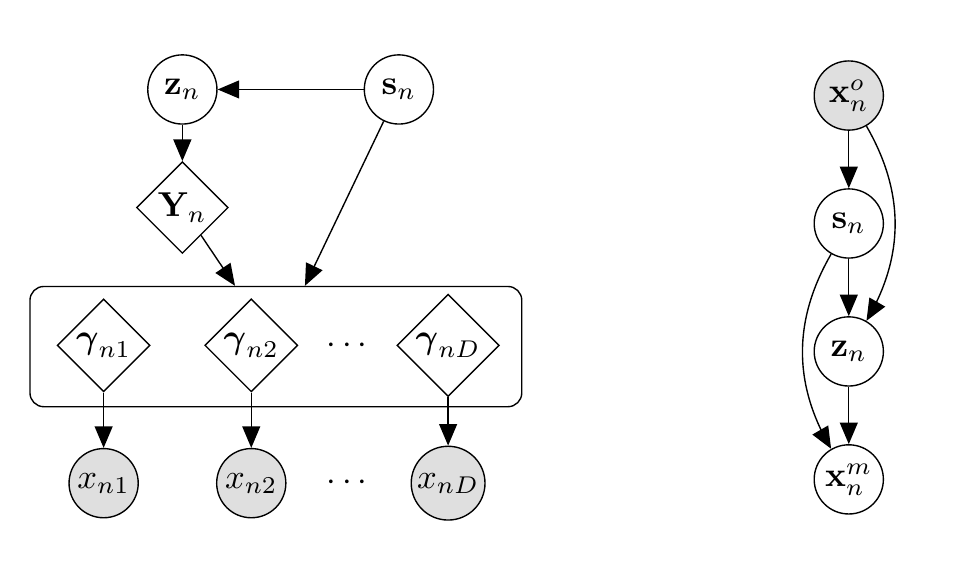}
\caption{Graphical models for the generative and recognition models of the HI-VAE.} \label{fig:recM}
\end{figure}

In order to prevent the KL term in~\eqref{eq:ELBO}  from dominating the  ELBO, thus penalizing rich posterior distributions for $\z_n$, we can impose structure in  the latent variable representation $\z_n$ through its prior distribution. 
We propose a Gaussian mixture prior $p(\z_n)$ \citep{Jang16}, such that
\begin{align}
p(\s_n) &= \text{Categorical}(\s_n|\boldsymbol{\pi}) \\ 
p(\z_n| \s_n) &= \cN(\z_n|\boldsymbol{\mu}_p(\s_n), \mathbf{I}_K),
\end{align} 
where $\s_n$ is a one-hot encoding vector representing the component  in the  mixture, i.e., the mean of the Gaussian component that generates $\z_n$. For simplicity, we assume a uniform Gaussian mixture with $\pi_\ell= 1/L$ for all $\ell$.  

Moreover, to allow the model to more accurately capture the statistical dependencies among heterogeneous attributes, we propose a hierarchical structure that allows different attributes to share network parameters (i.e., to amortize the generative model). More specifically,  we introduce an intermediate homogenous representation of the data $\Y =  [\y_{n1}, \ldots, \y_{nD}] $, which is jointly generated by a single DNN with input $\z_n$, $ \g(\z_n)$.  Then, the likelihood parameters of each attribute $d$ are the output of an independent DNN with inputs $\y_{nd}$ and $\s_n$, such that $p(x_{nd}| \boldsymbol{\gamma}_{nd}=\h_d(\y_{nd},\s_n))$.  
%
Note that, in this hierarchical structure, the top level (from $\z_n$ to $\Y_n$) captures statistical dependencies among the attributes through the shared DNN $\g(\z_n)$, while the bottom level in the hierarchy (from $\Y_n$ and $\s_n$ to $\x_n$) accounts for heterogeneous likelihood models using $d$ independent DNNs $\h_d(\y_{nd}, \s_n)$.
%
The resulting generative model, {that is hereafter referred to as Heterogeneous-Incomplete VAE (HI-VAE)}, is shown in Figure~\ref{fig:extensions} and is formulated as indicated in Table \ref{HIVAE-table}, 
which also shows how we parametrize in the HI-VAE the different likelihood models provided in Section 3.2.\footnote{Other likelihood functions (e.g., a Gamma distribution) and data types (e.g., interval data using e.g., a Beta distribution) can be readily be incorporated.}

Regarding the recognition network (Figure \ref{fig:recM})  $q(\s_n|\x_n^o)$ is a categorical distribution with a parameter vector $\boldsymbol{\pi}$ given by the output of a DNN with input $\tilde{\x}_n$ and a soft-max output function. Then, a concatenation of both $\s_n$ and $\tilde{\x}_n$ is used to construct the moments of the Gaussian $q(\z_n|\x_n^o, \s_n)$ posterior distribution via two independent NNs. Finally, to enforce that the model captures all correlations using the hidden variables $\s_n$ and $\z_n$, in the recognition network we assume that the posterior distribution of the missing attributes $\x_n^ m$ is conditionally independent on the observed attributes $\x_n^o$, given $\s_n$ and $\z_n$. 
%
Hence, our variational distribution (or, equivalently, our recognition model) factorizes as:
$$q(\s_n,\z_n, \x_n^m|\x_n^o ) = q(\s_n| \x_n^o) q(\z_n | \x_n^o, \s_n ) \prod_{d\in \Mn} p(x_{nd} | \z_n, \s_n ).$$
By combining the HI-VAE generative model and the proposed recognition network, we derive the expression for ELBO in Table \ref{HIVAE-table}, where the Gumbel-softmax reparameterization trick \citep{Jang16} is used to draw differentiable samples from $q(\s_n, \z_n |\x_n^o )$.


\section{Experiments}\label{sec:exp}
%
%
%
%
%
%
In this section,  we first evaluate the performance of the HI-VAE  at solving a  missing data imputation task  in heterogeneous data, comparing it to other methods in the literature. Then, we focus on a classification task, where we evaluate the classification degradation due to performing mean imputation for the missing data in supervised models compared to using the fully generative HI-VAE, which does not require data imputation. The models and datasets employed in the experiments can be found in the following public repository {\footnotesize \url{https://github.com/probabilistic-learning/HI-VAE}}.

%

\subsection{Missing data imputation}
%
In our first experiment, we evaluate the performance of the proposed HI-VAE at imputing missing data.
We use six different heterogenous datasets from the UCI repository~\citep{Lichman}, which vary both in the number of instances and attributes, as well as in the statistical data types of the attributes. We summarize the main characteristics of these databases in Table \ref{Tabladatasets}. For each dataset we generate $10$ different incomplete datasets, removing completely at random a percentage of the data ranging from a $10\%$ deletion to a $50\%$.

\begin{table}[!]
\caption{\emph{{Dataset description. The attributes include the target variable for those datasets that have an associated binary classification task.}}} \label{Tabladatasets}
\begin{tabular}{llllllll}
Database & Objects & Attributes & \# Real & \# Positive & \# Categorical & \# Ordinal & \# Count\\
\hline
Adult & 32561 & 12 & 0 & 3 & 6 & 1 & 2\\
Breast & 699 & 10 & 0 & 0 & 1 & 9 & 0\\
Default Credit & 30000 & 24 & 6 & 7 & 4 & 6 & 1\\
Letter & 20000 & 17 & 0 & 0 & 1 & 16 & 0\\
Spam & 4601 & 58 & 0 & 57 & 1 & 0 & 0\\
Wine & 6497 & 13 & 0 & 11 & 1 & 0 & 1\\
\hline
\end{tabular}
\end{table}

\vspace{5mm}
\xhdr{Imputation strategy}
Once the HI-VAE model is trained, the imputation of missing data is performed in a three-step process: First, we perform the MAP estimate of $q(\z_n,\s_n|\x_n^o)$ to obtain $\hat{\z}_n$ and $\hat{\s}_n$. With these MAP estimates, we evaluate the generative model, obtaining $\hat{\Y}_n = \g(\hat{\z}_n)$ and $\hat{\gamma}_{nd}=\h_d(\hat{\y}_{nd},\hat{\s}_n)$ for every attribute. Finally, the imputed values $\hat{\x}_n$ are obtained as the mode of each distribution $p(x_{nd}|\hat{\gamma}_{nd})$, where the computation of the mode depends on the likelihood model of the attribute. A further discussion on imputation methods is  provided in Section \ref{exp_sub3}.

\vspace{5mm}
\xhdr{Imputation error}
We compare the above models in terms of average imputation error computed as $\text{AvgErr} = 1/D \sum_d \text{err}(d)$, where we use the following error metrics for each attribute, since the computation of the errors depends on the type of variable we are considering:
\begin{itemize}
\item Normalized Root Mean Square Error (NRMSE) for numerical variables, i.e.,
\begin{equation}
\text{err}(d) = \frac{\sqrt{1/n \sum_n (x_{nd}- \hat{x}_{nd})^2}}{\max(\x_{d})- \min(\x_{d})}.
\end{equation}
\item Accuracy error for categorical variables, i.e.,
\begin{equation}
\text{err}(d) ={\frac{1}{n} \sum_n I(x_{nd} \neq \hat{x}_{nd} ) } .
\end{equation}
\item Displacement error  for ordinal variables, i.e.,
\begin{equation}
\text{err}(d) ={\frac{1}{n} \sum_n | \frac{x_{nd} - \hat{x}_{nd}}{R} } |.
\end{equation}
\end{itemize}

\vspace{5mm}
\xhdr{Comparison}
We compare the performance of the following methods for missing data imputation:
\begin{compactitem}
\item \textbf{Mean Imputation}: We use as baseline an algorithm that imputes the mean of each continuous attribute and the mode of each discrete attribute.
\item \textbf{MICE:}~ Multiple Imputation by Chained Equations~\citep{MICE11}, which is an iterative method that performs a series of supervised regression models, in which missing data is modeled conditional upon the other variables in the data, including those imputed in previous rounds of the algorithm. We use  MICE implementation within the  \emph{fancyimpute} package {\footnotesize \url{https://github.com/iskandr/fancyimpute}}, which, in its current implementation, only allows the user to pick a homogeneous regression model for all attributes, independently of whether they are numerical or nominal. 
\item \textbf{GLFM:} General latent feature model for heterogeneous data ~\citep{valera2017general}, which was initially introduced for table completion in heterogeneous datasets in ~\citep{Valera14}. This method handles all the numerical and nominal data types described in Section~\ref{sec:model} and performs MCMC inference. We run 5000 iterations of the sampler using the available implementation in {\footnotesize \url{https://github.com/ivaleraM/GLFM}}. 
\item \textbf{GAIN:} Generative adversarial network for missing data imputation~\citep{yoon2018gain}, which  uses MSE as a loss function for numerical variables, and cross-entropy for binary variables.  We train GAIN for $2000$ epochs using the network specifications and hyperparameters  reported in \citep{yoon2018gain}.
\item \textbf{HI-VAE:} Model introduced in Section~\ref{extensions}, which we implement in TensorFlow using only one dense layer for all the parameters of the encoder and decoder of the HI-VAE). We set the dimensionality of $\z,\y$ and $\s$ to 10, 5 and 10, respectively. The parameter $\tau$ of the Gumbel-Softmax is annealed using a linear decreasing function on the number of epochs, from $1$ to $10^{-3}$. 
We train our algorithms for $2000$ epochs using minibatches of 1000 samples. {We note that we have used the same NN architecture in all experiments and, therefore,  further improvements could be achieved by cross-validating the architecture for each database. We further explore this aspect in Section \ref{exp_sub3}.}
\end{compactitem}

%
%
%

\subsubsection{Variations in the HI-VAE design}\label{exp_sub3}

First, we explore different aspects of the design (such as network architecture, normalization layer, and hyperparameter selection) and how the use of the missing data imputation strategy of HI-VAE may improve the performance of the proposed HI-VAE for missing data estimation.

\vspace{5mm}
\xhdr{Network design}    
Here we analyze the sensitivity of the HI-VAE to the network architecture. To this end, we vary dimensionality of $\s$, $\z$ and $\y$ and 
consider both generator and inference networks with either one or two  dense layers with ReLu activation functions.
In Figure \ref{barplot_configs} we show the HI-VAE average imputation error with a 20\% missing data rate for different network configurations and latent space dimensions.  Here we observe that, while using two layers and a larger latent dimension (brown bars) tend to improve the performance, significant gains are only observed with Letter database, where more complex architectures lead to a lower imputation error. 

\begin{figure*}[ht]
\centering
\centering\includegraphics[scale=1.0]{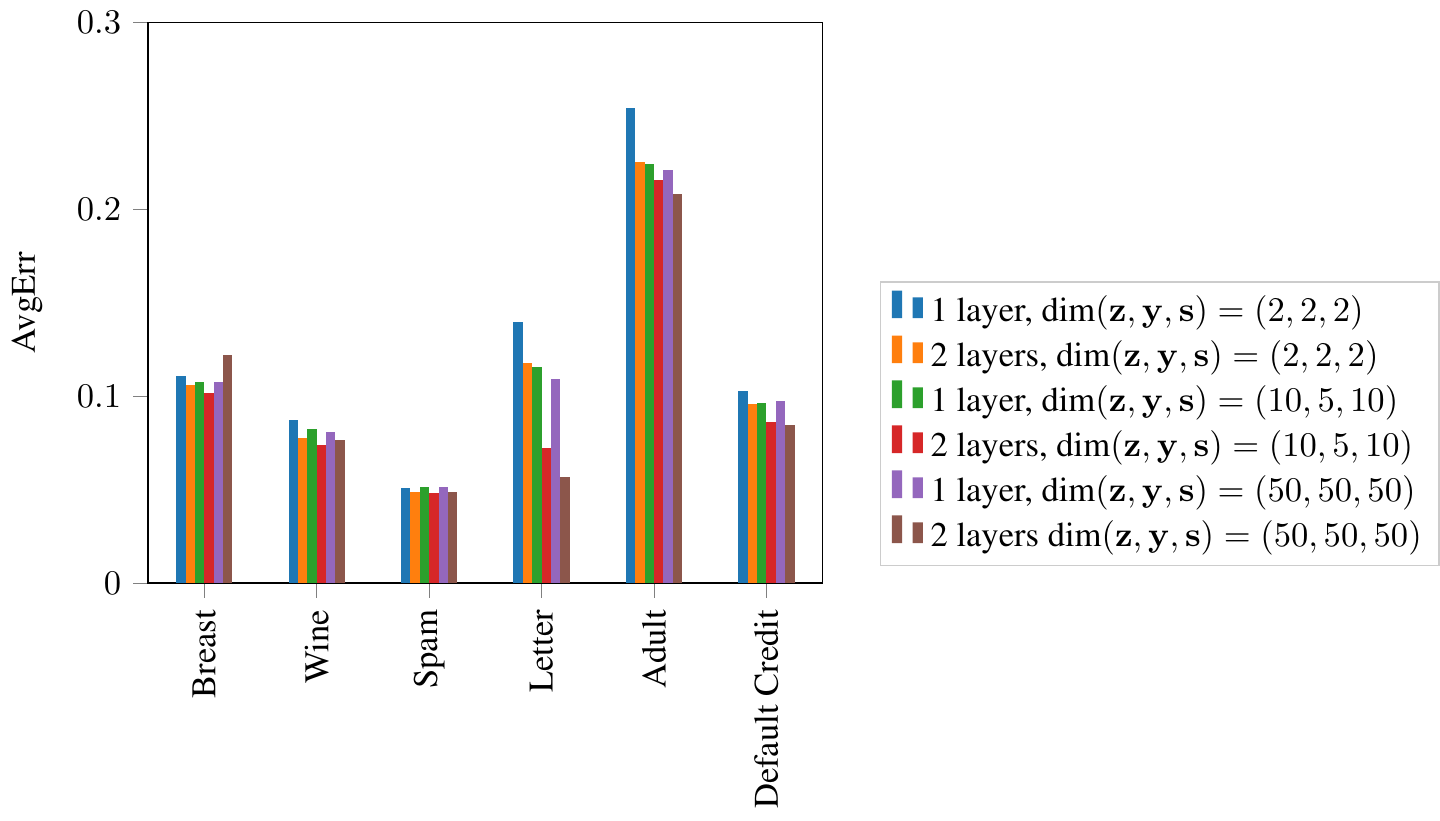}
\caption{HI-VAE average imputation error for different network configurations and latent space dimensions with a 20\% rate of missing data.} \label{barplot_configs}
\end{figure*}

\xhdr{HI-VAE imputation strategy}
Once we have trained the generative model, to impute missing data we can either sample from the generative model or use the inferred parameters of the output distribution, e.g., impute the mode of the inferred distribution.  To illustrate the differences, 
we show in Figure~\ref{fig:visual1} and \ref{fig:visual2} the goodness of fit provided by the HI-VAE and the GLFM in a positive real-valued variable and a categorical variable with 6 categories, both belonging to the Adult dataset.
Specifically, we show (top row) the true distribution of the data together with the HI-VAE output distribution for the observed data and the HI-VAE output distribution for the missing values. We show results for HI-VAE using the mode of the distribution and HI-VAE using one sample for imputation. We also show results for the GLFM. Further, in the bottom row we show the Q-Q plot for the positive-real variable and the confusion matrix for the categorical variable (see the figure caption for more details). Note that, while both the HI-VAE and the GLFM result in a good fit of the positive real variable (although the HI-VAE provides a smoother, and thus, more realistic distribution for the data); the GLFM fails at capturing the categorical variable--it assigns all the probability to a single category. These results are consistent with Table \ref{nominal} in the paper, which demonstrate the superior ability of the HI-VAE to perform missing data imputation in nominal variables. 

\begin{figure*}[ht]
\centering\includegraphics[scale=0.5]{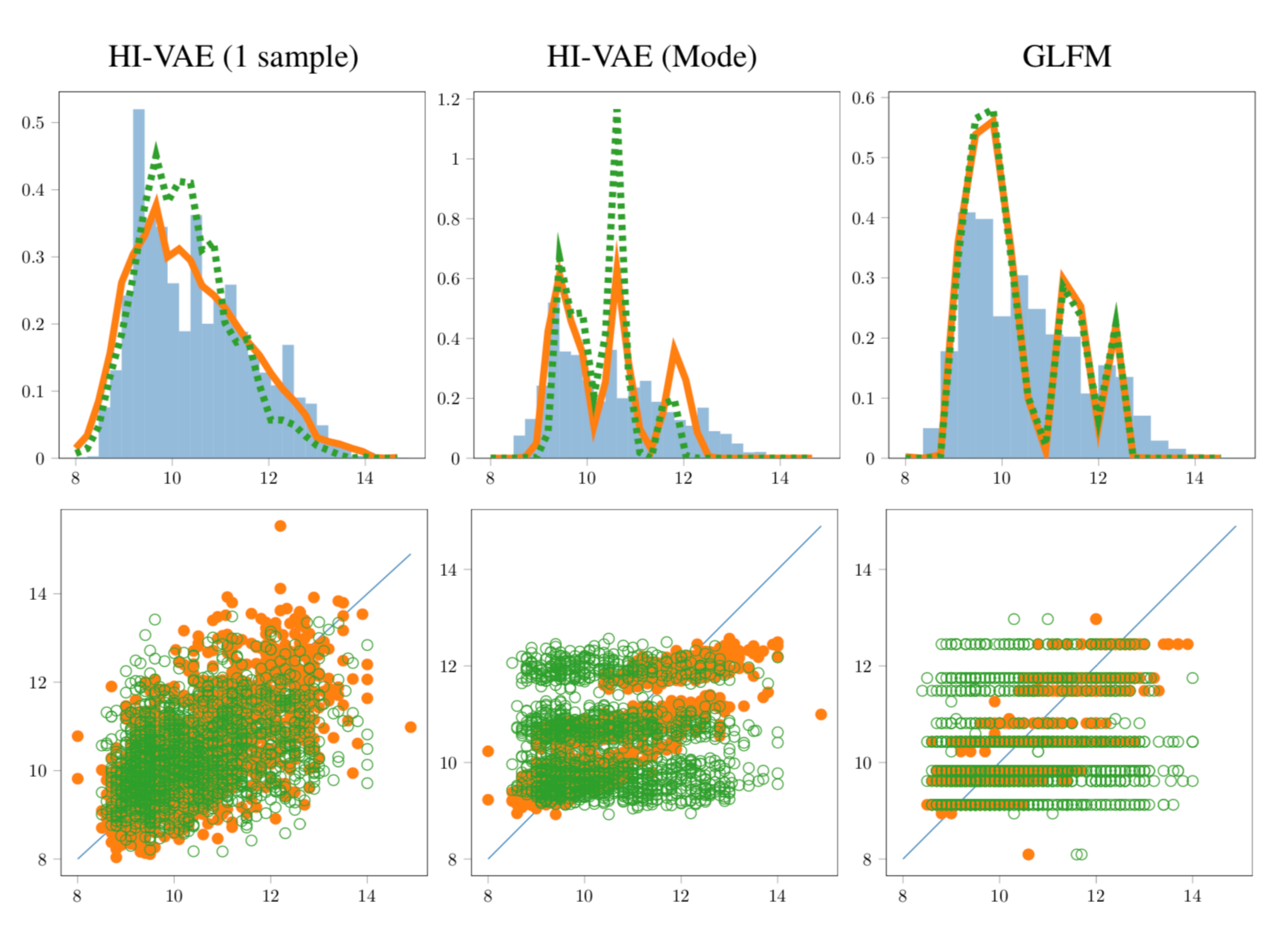}
\caption{We demonstrate the fit provided by the HI-VAE and the GLFM in a positive real-valued variable  of the Adult dataset. Top row depicts the true empirical data distribution (shadowed histogram) and the inferred data distribution for the observed attributes in dashed line and for the missing data in solid line. The bottom row shows the Q-Q plot (observed in orange ($\bullet$) marker  and missing in green ($\circ$) marker). The left-most column shows the results for the HI-VAE when we sample from the model posterior distribution (given the observed data) to impute, while for the center column we use the mode of the posterior. }\label{fig:visual1}
    \end{figure*}
    \begin{figure*}[ht]
\includegraphics[scale=0.4]{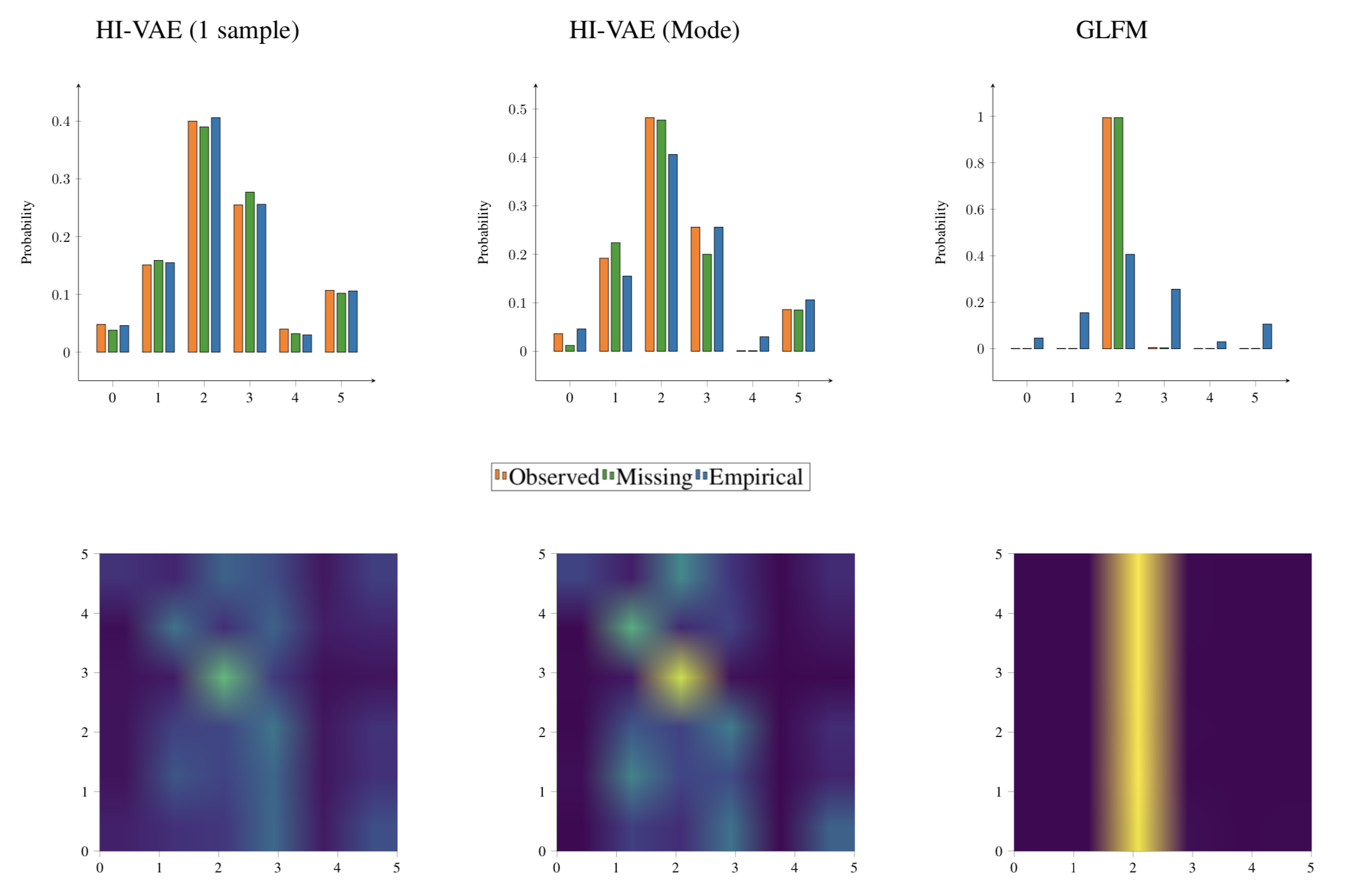}
\caption{We demonstrate the fit provided by the HI-VAE and the GLFM in a categorical variable with 6 categories  of the Adult dataset. Top row depicts the true empirical data distribution and the inferred data distribution for the observed attributes and for the missing data. The bottom row shows the missing data confusion matrix.}\label{fig:visual2}
\end{figure*}

\vspace{5mm}
\xhdr{Normalization layer}     
Finally, we study the effect of the  batch (de-)normalization layer at the input of the reconstruction (and at the output layer of the generative) model for the numerical variables described in Section~\ref{sec:het}. The first two rows of Tables~\ref{numeric}~and~\ref{nominal} show the imputation error of the HI-VAE with and without the (de-)normalization layers for all the considered datasets containing numerical variables (since the normalization layer only apply to numerical variables), and for a 20\% of missing data selected completely at random.  
Here we observe that the normalization layer  not only leads to a significant improvement in terms of imputation error for the Adult, the Spam and the Wine datasets, but it also prevents numerical errors from occurring during inference -- for the Default dataset, the gradients of the ELBO with respect to the model parameters take infinite values. 


%
%


\subsubsection{Comparison with exiting methods}
\begin{table}[h]
\caption{\emph{Imputation error.} Average and standard deviation of the imputation error for a 20\% of missing data,  evaluated exclusively over \textbf{numeric variables}. }\label{numeric}
\begin{tabular}{lllllll}
Model           & Adult             & Breast        & DefaultCredit     & Letter        & Spam              & Wine              \\
\hline
HI-VAE (no norm.)          & $0.210 \pm 0.028$ & $-$ & $Inf$ & $-$ & $0.054 \pm 0.018$ & $ 0.165 \pm 0.042$ \\
 HI-VAE           & $0.106 \pm 0.002$ & $-$ & $0.043 \pm 0.001$ & $-$ & $0.052 \pm 0.001$ & $\bf 0.074 \pm 0.001$ \\
\hline
 Mean imputation & $0.111 \pm 0.002$ & $-$ & $0.056 \pm 0.001$ & $-$ & $0.053 \pm 0.001$ & $0.103 \pm 0.002$ \\
 MICE            & $0.108 \pm 0.002$ & $-$ & $\bf 0.035 \pm 0.002$ & $-$ & $0.052 \pm 0.003$ & $0.074 \pm 0.002$ \\
 GLFM            & $\bf 0.083 \pm 0.001$ & $-$ & $0.051 \pm 0.005$ & $-$ & $0.052 \pm 0.001$ & $0.082 \pm 0.004$ \\
 GAIN            & $0.225 \pm 0.192$ & $-$ & $0.044 \pm 0.002$ & $-$ & $\bf 0.049 \pm 0.001$ & $0.086 \pm 0.002$ \\
\hline
\end{tabular}
\end{table}

\begin{table*}[t!]
\caption{\emph{Imputation error.} Average and standard deviation of the imputation error for a 20\% of missing data, evaluated exclusively over \textbf{nominal variables}. }\label{nominal}
\hspace{-2cm}\begin{tabular}{lllllll}
 Model           & Adult             & Breast            & DefaultCredit     & Letter            & Spam              & Wine              \\
\hline
  HI-VAE (no norm.)              & $0.406 \pm 0.005$ & $-$ & $0.202 \pm 0.003$ & $-$ & $0.166 \pm 0.019$ & $0.245 \pm 0.017$ \\
 HI-VAE           & $\bf 0.304 \pm 0.006$ & $0.112 \pm 0.003$ & $\bf 0.158 \pm 0.001$ & $\bf 0.105 \pm 0.002$ & $\bf 0.111 \pm 0.009$ & $0.016 \pm 0.003$ \\
 \hline
 Mean imputation & $0.405 \pm 0.002$ & $0.211 \pm 0.006$ & $0.2 \pm 0.001$   & $0.162 \pm 0.002$ & $0.393 \pm 0.014$ & $0.248 \pm 0.014$ \\
 MICE            & $0.601 \pm 0.002$ & $0.111 \pm 0.002$ & $0.163 \pm 0.003$ & $0.133 \pm 0.0$   & $0.168 \pm 0.012$ & $0.02 \pm 0.004$  \\
 GLFM            & $0.407 \pm 0.003$ & $\bf 0.076 \pm 0.003$ & $0.236 \pm 0.012$ & $0.161 \pm 0.001$ & $0.154 \pm 0.02$  & $\bf 0.006 \pm 0.001$ \\
 GAIN            & $0.66 \pm 0.025$  & $0.16 \pm 0.009$  & $0.211 \pm 0.005$ & $0.164 \pm 0.001$ & $0.276 \pm 0.017$ & $0.236 \pm 0.014$ \\
\hline
\end{tabular}
\end{table*}

Finally, we compare the performance of the HI-VAE with existing methods in the literature to input missing data in heterogenous datasets. 
For the  HI-VAE we use here a relatively-simple configuration with a single dense layer with a latent space of dimensions $\text{dim}(\z)=10, \text{dim}(\s)=10$ and $\text{dim}(\y)=10$. A more careful design of the HI-VAE structural parameters for each dataset may thus improve the HI-VAE performance. 
Figure~\ref{fig:missing} summarizes the average imputation error for each database as we vary the fraction of missing data. {The results clearly show that the proposed HI-VAE is the only method that consistently outperforms mean imputation in all the datasets---since mean imputation assumes all the attributes to be independent, any missing data imputation method that accounts for statistical dependencies in the data should perform at least as accurately as mean imputation. }The second more robust model is the GLFM, which performs best in small datasets (Breast and Wine). This might be explained by the fact that, while it accounts for mixed nominal and discrete data, it relies on Gibbs-sampling for inference, scaling and mixing poorly for larger datasets. In contrast, the MICE and GAIN\footnote{We would like to clarify that the reported results do not quite match those provided in~\citep{yoon2018gain}, despite using the code  and the hyperparameters provided by the authors. For the sake of reproducibility, we will incorporate the GAIN implementation  to our public repository.} are  outperformed by the Mean-imputation baseline in several datasets, most likely, due to the fact that they do not account for different types of mixed nominal and numerical attributes. 


\begin{figure*}[t!]
\centering\includegraphics[scale=0.7]{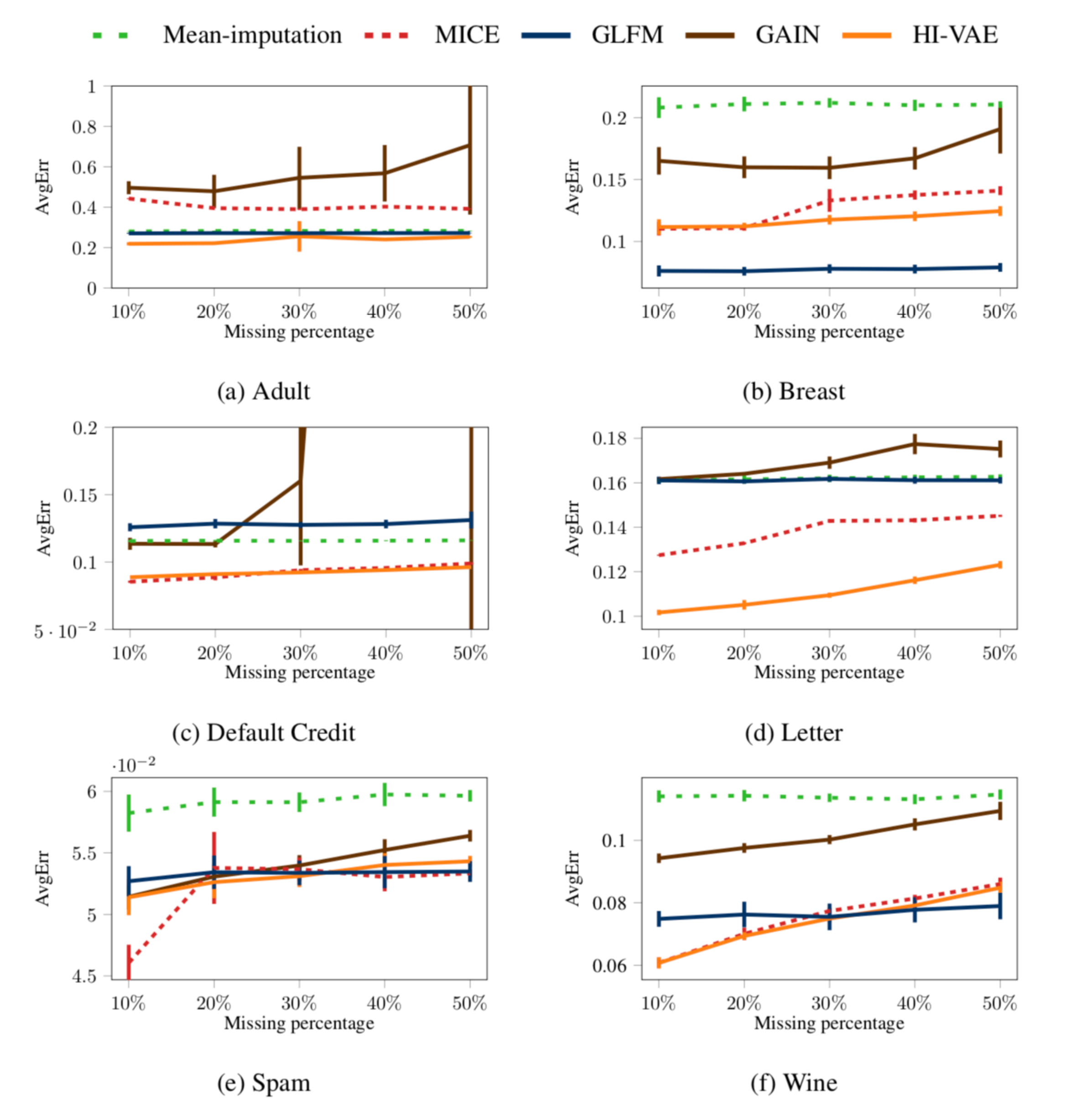}
\caption{\emph{Missing Data.} Average imputation error for different percentages of missing data (completely at random).
}\label{fig:missing}
\end{figure*}

A deeper understanding of the results in Figure~\ref{fig:missing} can be obtained by separately analyzing the error in numeric variables (real, positive and count variables) in Table \ref{numeric}, and nominal variables (categorical/ordinal variables) in Table \ref{nominal}. In both cases, we use 20\% missing data. While for numeric variables HI-VAE achieves a comparable error w.r.t. the rest of the methods, it is in the imputation of nominal variables where HI-VAE achieves a remarkable gain, being the best performing method in four out of six cases. These results demonstrate the superior ability of HI-VAE to exploit underlying correlations among the set of heterogeneous attributes. 
%
%

%
%
%
%
%

\subsection{Predictive Task}
Although the HI-VAE is a fully unsupervised generative model, we evaluate its performance at solving a classification task, a multi-class classification problem for the Letter dataset (with 26 classes corresponding to the different letters) and a binary classification problem for the rest of databases (predicting  {the binary label of each element of the dataset}).
%
{The idea behind this experiment, is to treat the classes to be predicted as missing entries in the target attribute, using HI-VAE to provide an imputation of these missing entries.}
We use 50\% of the data for training, which for HI-VAE means that we remove 50\% of the labels {in the target attribute} to train the generative model.  
Regarding the training data, we consider three different scenarios: the first assumes complete input attributes in the training set (no missing data), the second assumes 10\% missing values in the input training data, and the third assumes 50\% missing values. Since the supervised methods we compare HI-VAE to cannot handle missing data, we impute the mean (or the mode for discrete attributes) of each  attribute to the missing input values during training.
Here, we compare our HI-VAE with two supervised methods: deep logistic regression (DLR) and  the conditional VAE (CVAE) in \citep{Kiyuk15}.  
Following our results in Figure \ref{barplot_configs}, we use the basic configuration for the HI-VAE, i.e., one dense layer and $\z,\y$ and $\s$ to 10, 5 and 10, respectively, for all datasets except for the Letter, for which we  use two dense layers with ReLU activations and 50-dimensional latent spaces. 

\xhdr{Results}
Table~\ref{tab:pred} summarizes the results, where we observe that our HI-VAE method provides competitive results in all cases.  
%
%
%
  Furthermore, note HI-VAE provides the best results for both Wine and Breast, while showing less degradation with increasing fraction of missing input data in the DefaultCredit and Spam. 
These results show that a fully generative model might be preferred over a supervised model with imputed data. 

\begin{table*}[]
\caption{\emph{Prediction Accuracy.} Average and standard deviation of the classification error when we use 50\% of the labels for training and assume complete input data and 10\% and 50\% of missing values in input data (right-hand table).} 
\label{tab:pred}
\hspace{-1cm}\begin{tabular}{cllllll}
\hline
 \% Missing             & Model   & Breast            & DefaultCredit     & Letter            & Spam              & Wine              \\
\hline
 \multirow{3}{*}{0\%}  & DLR     & $0.041 \pm 0.01$  & $\bf 0.179 \pm 0.002$ & $0.142 \pm 0.003$ & $\bf 0.081 \pm 0.005$ & $0.018 \pm 0.003$ \\
                       & CVAE    & $0.04 \pm 0.012$  & $\bf 0.179 \pm 0.001$ & $0.14 \pm 0.004$  & $0.081 \pm 0.006$ & $0.016 \pm 0.002$ \\
                       & HIVAE   & $\bf 0.026 \pm 0.005$ & $0.2 \pm 0.004$   & $\bf 0.117 \pm 0.014$ & $0.096 \pm 0.007$ & $\bf 0.014 \pm 0.002$ \\
                       \hline
 \multirow{3}{*}{10\%} & DLR     & $0.04 \pm 0.009$  & $\bf 0.184 \pm 0.001$ & $0.229 \pm 0.002$ & $0.09 \pm 0.005$  & $0.027 \pm 0.003$ \\
                       & CVAE    & $0.048 \pm 0.009$ & $0.184 \pm 0.002$ & $0.227 \pm 0.003$ & $\bf 0.088 \pm 0.006$ & $0.025 \pm 0.003$ \\
                       & HIVAE   & $\bf 0.031 \pm 0.007$ & $0.201 \pm 0.002$ & $\bf 0.212 \pm 0.017$ & $0.103 \pm 0.008$ & $\bf 0.022 \pm 0.006$ \\
                       \hline
 \multirow{3}{*}{50\%} & DLR     & $0.08 \pm 0.014$  & $\bf 0.196 \pm 0.003$ & $\bf 0.496 \pm 0.005$ & $\bf 0.134 \pm 0.008$ & $0.078 \pm 0.006$ \\
                       & CVAE    & $0.101 \pm 0.038$ & $0.197 \pm 0.003$ & $\bf 0.496 \pm 0.005$ & $0.138 \pm 0.009$ & $0.078 \pm 0.005$ \\
                       & HIVAE   & $\bf 0.052 \pm 0.012$ & $0.205 \pm 0.003$ & $0.589 \pm 0.014$ & $0.138 \pm 0.005$ & $\bf 0.042 \pm 0.005$ \\
\hline
\end{tabular}
\end{table*}

\section{Conclusions}

In this paper, we focus on designing and inferring deep generative models (in particular, VAEs) for heterogeneous and incomplete data. We note that it is not a  straightforward problem, and that it has been overlooked in the literature. The main issues covered in this paper and for which HI-VAE provides an effective solution can be summarized as follow: First, standard (and conditional) VAEs assume complete data during training, however, missing data imputation is a fully unsupervised task where missing values may appear ubiquitously in the dataset. Unfortunately, while VAEs perform accurate inference through a recognition model sharing parameters among inputs, this is not directly possible when training data is incomplete (DNNs require complete input). 
In HI-VAE, we derive a lower-bound on the data marginal likelihood that depends exclusively on the observed data. Also, we propose methods to deal with missing values in the recognition network. Second, when data are heterogeneous in both statistical types and ranges, the inference of a joint set of parameters that accurately captures the statistical dependencies among attributes results in a complex optimization problem with many local optima. Intuitively, each local optima potentially captures the correlations between a subset of attributes and treats the rest as independent, while the global optima captures all the existing correlations in the data. In HI-VAE, we enforce correlation by using a joint DNN to construct the parameters that define the output distribution of each of the attributes. Third, in contrast to deep generative approaches for structured and homogeneous data (e.g., images or text), the use of more complex DNNs (e.g., CNNs or RNNs) does not necessarily lead to a better fitting of the data in heterogeneous datasets, where there is no clear notion of correlation to be exploited by weight sharing of the DNNs. The hierarchical HI-VAE generative model captures correlation among the different attributes by using a latent space spanned by a Gaussian mixture.

Our empirical results show that our proposed HI-VAE outperforms competitors on a heterogenous data completion task and provides comparable results in classification accuracy to deep supervised methods, which cannot handle missing values in the input data, therefore, requiring imputation of missing inputs in the data. Future work includes the extension to more complex attributes such as images or text, and the generalization to temporal heterogeneous series with missing data.

%

\section{Acknowledgments}

The authors wish to thank Christopher K. I. Williams, for fruitful discussions and helpful comments to the manuscript. Alfredo Nazabal would like to acknowledge the funding provided by the UK Government’s Defence \& Security Programme in support of the Alan Turing Institute. The work of Pablo M. Olmos is sup-ported by Spanish government MCI under grant PID2019-108539RB-C22 and RTI2018-099655-B-100, by Comunidad de Madrid under grants IND2017/TIC-7618, IND2018/TIC-9649, and Y2018/TCS-4705, by BBVA Foundation under the Deep-DARWiN project, and by the European Union (FEDER and the European ResearchCouncil  (ERC)  through  the  European  Unions  Horizon  2020  research  andinnovation program under Grant 714161).  
Zoubin Ghahramani acknowledges support from the Alan Turing Institute (EPSRC Grant EP/N510129/1) and EPSRC Grant EP/N014162/1,  and  donations  from  Google  and  Microsoft Research.  Isabel Valera is supported by the MPG Minerva Fast Track program. We  also  gratefully  acknowledge the support of NVIDIA Corporation with the donation of the Titan X Pascal GPU used for this research.

\section*{References}
\bibliography{Bib}

\end{document}